# Federated Knowledge Graph Unlearning via Diffusion Model

Bingchen Liu, Yuanyuan Fang

*Abstract*—Federated learning (FL) promotes the development and application of artificial intelligence technologies by enabling model sharing and collaboration while safeguarding data privacy. Knowledge graph (KG) embedding representation provides a foundation for knowledge reasoning and applications by mapping entities and relations into vector space. Federated KG embedding enables the utilization of knowledge from diverse client sources while safeguarding the privacy of local data. However, due to demands such as privacy protection and the need to adapt to dynamic data changes, investigations into machine unlearning (MU) have been sparked. However, it is challenging to maintain the performance of KG embedding models while forgetting the influence of specific forgotten data on the model. In this paper, we propose FedDM, a novel framework tailored for machine unlearning in federated knowledge graphs. Leveraging diffusion models, we generate noisy data to sensibly mitigate the influence of specific knowledge on FL models while preserving the overall performance concerning the remaining data. We conduct experimental evaluations on benchmark datasets to assess the efficacy of the proposed model. Extensive experiments demonstrate that FedDM yields promising results in knowledge forgetting.

## I. INTRODUCTION

FEDERATED learning (FL) is a decentralized learning approach that enables collaborative model training across multiple geographically distributed devices or data sources while preserving data privacy. It achieves model sharing and collaboration across devices by training models locally on each device and then aggregating the updated parameters. FL has been widely applied in various fields such as news recommendation [1] and healthcare [2], offering solutions to challenges like data privacy and resource decentralization. Knowledge graph (KG) is a structured representation of knowledge, organizing information into entities and relationships to facilitate understanding and reasoning. KG embedding maps entities and relationships into vector space, effectively encoding structural and semantic information for link prediction [3] and other downstream applications [4]. Federated KG embedding learning [5] [6] [7] extends this approach to decentralized environments, enabling collaborative model training across distributed data sources while preserving data privacy.

(Corresponding author: Bingchen Liu)
Bingchen Liu is with the School of Software, Shandong University, Jinan, 250098, China.(e-mail: lbcraf2018@126.com)
Yuanyuan Fang is with the School of Computer Science and Technology, Shandong University of Finance and Economics, Jinan, 250220, China.

Federated KG embedding represents a vast amount of knowledge from the real world, often facing two main challenges: Firstly, knowledge represented by KGs may become outdated or erroneous over time, necessitating the elimination of such data's impact on model performance. Secondly, the enactment of privacy laws like the General Data Protection Regulation (GDPR) in the European Union and the California Consumer Privacy Act (CCPA) requires the right to be forgotten, necessitating timely removal of requested privacy data from training datasets and eliminating its impact on trained models. While it's relatively easy to remove forgotten samples from KG training datasets, their impact on already trained models is often challenging to eliminate. Retraining on the data after removing forgotten samples can be time and resource-intensive. Moreover, assessing the specific impact of forgotten knowledge on the overall model is challenging due to the correlations between knowledge and the randomness of training.

Therefore, Machine unlearning (MU) in the context of federated KG embedding is conceived to enable the complete and rapid removal of samples and their impacts from training datasets and federated KG embedding models, while preserving the overall performance of the model. For example, an example of applying federated KG to an e-commerce website that needs to implement machine unlearning is illustrated in Figure 1. Ferdinand and Sophia, due to the fact that they have purchased valuable items such as gold necklaces and gold rings, may propose to remove the memory of the dataset and the trained model for the interactions with this part of the items due to the need to protect their privacy. An existing work [8] based on the foundation of cognitive neuroscience, combines retroactive interference and passive decay to achieve knowledge forgetting in federated KG embedding models, yet they overlook the exploration of using generative models for MU.

To address the above issues, we propose FedDM, a new diffusion model-based framework for performing MU on federated KG embeddings. Specifically, we use the diffusion model to learn to generate embedding representations of forgotten knowledge, and utilize the similarly noisy data generated after learning to replace the original embedding representations of forgotten knowledge. The noise property of the diffusion generation model is fully utilized to dilute the characteristics of the representation of the data that needs to be forgotten. Thus, while ensuring the overall performance of the model, we eliminate the influence of specific knowledge on the model as a whole in the federal KG embedding framework to achieve the purpose of knowledge forgetting.

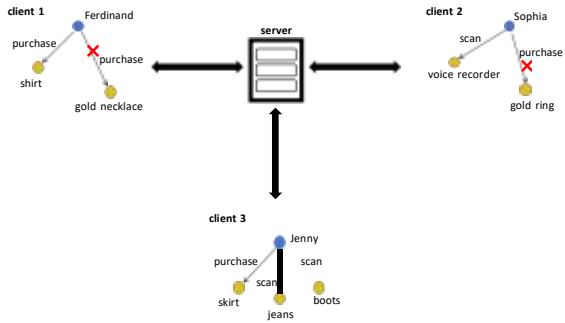

Fig. 1: An example of a federated knowledge graph requiring machine unlearning.

In summary, our contributions are listed as below.

. We propose FedDM, a new diffusion model-based framework for MU of federated KG embeddings. The framework makes full use of the noise property of the diffusion model to dilute the representation features of forgotten data. It not only eliminates the impact of knowledge that needs to be forgotten on the model as a whole, but also maintains the overall performance of the federal KG embedding framework.

. We demonstrate the effectiveness of our approach on newly-constructed datasets. Extensive experimental results show that our method can achieve advanced results.

The remaining parts of this paper are organized as follows. Related work is summarized in section II. The problem is formulated and defined in section III. Then, we introduce our proposed model in section IV. After that, we conduct experiments and discuss the experimental results in section V. At last, section VI gives the conclusion of this paper.

## II. RELATED WORK

### A. Knowledge Graph Embedding

KG embedding involves mapping entities and relationships in a knowledge graph into low-dimensional vector spaces, facilitating efficient representation learning for KGs. Khobragade et al. [9] combine a simplified graph attention mechanism with a neural network to learn KG embeddings without negative sampling requirements. Buosi et al. [10] use knowledge graph embedding to train models in order to predict recurrence in early stage non-small cell lung cancer patients, which can be used as an effective complement in classification systems and contribute to the treatment of cancer patients. Jiang et al. [11] perform link prediction of knowledge graphs based on KG embedding of multi-source hierarchical neural networks to cope with the heterogeneity of KG entities and relationships, and effectively extract complex graph information. The study of federated KG embedding has attracted attention due to privacy protection and other needs. Hu et al [5] conducted the first comprehensive study of privacy threats in emerging federated KG embedding from attack and defence perspectives. Compared to traditional KG embedding research, federated KG embedding is trained collaboratively from distributed KGs held between clients, avoiding the leakage and interaction of sensitive information in the original KGs of clients.

### B. Federated Learning

Federated learning enables collaborative model training across decentralized devices or data sources while preserving data privacy. Wu et al [12] apply an efficient adaptive algorithm to FL for solving the training convergence problem during training and demonstrated that a simple combination of FL and adaptive methods can lead to divergence by providing a counterexample. To address the problem of statistical heterogeneity that weakens the client-side generalization ability of the global model, Zhang et al [13] use a joint learning approach with adaptive local aggregation to capture the required information in the global model of the client model in personalized FL. In addition to the FL research in the above areas, federated KG learning is also being focused on by an increasing number of researchers. In order to effectively solve the drift problem between local optimization and global convergence due to data heterogeneity, Zhu et al [8] propose a mutual knowledge distillation method to transfer local knowledge to the global and absorb knowledge to the global. Because of the requirements of privacy protection and data timeliness, the problem of MU in the federal KG domain needs to be drawn attention.

### C. Machine Unlearning

Machine unlearning refers to the process of removing specific data samples and their impacts from a trained model, often to comply with privacy regulations or to adapt to changing data distributions. Existing MU methods can be categorized into two mainstream ideas, one is data reorganization and the other is model manipulation. In model manipulation, the model provider achieves MU by tuning the parameters of the model. For example, Golatkar et al [14] propose a new hybrid privacy based forgetting scheme where the training dataset can be divided into core and user data, where the model training on the core data is non-convex and further training on the user data based on a quadratic loss function is performed to meet the requirements of a specific user task. Data reorganization is a technique where the model provider corrupts the data by reorganizing the training dataset. Zhang et al [15] use generative models in the field of image retrieval to create noisy data for data reorganization in order to adjust the weight of the retrieval model for the purpose of forgetting. However, there is no relevant method of generative modeling applied to the federated KG embedding domain for implementing MU. According to a new survey [16] points out that MU in the federated learning perspective is a direction worth exploring. However, there is no research exploring the idea of generative modeling for implementing MU in the federated KG domain. Inspired by [15], we explored the effectiveness of diffusion-based modeling to solve the MU problem in the context of federated KG embedding.

## III. PROBLEM FORMULATION

The federated KG framework consists of a central server and n clients, where the global embedding representation $E_g$

of entities is stored in the central server and each client stores embedding representation $E_l$ of its own local entities. The local KG in each client is divided into two parts: one part is the forget set, which represents the real-world dataset that needs to be removed due to requirements such as privacy or data timeliness; and the other is the retrained set, which represents the the remaining data in the full set of data after the removal of the forget set.

The problem of implementing machine unlearning in federated KG can be shown as follows: first, the central server updates the entity embeddings $E_l$ of each client batch by batch through entity embeddings $E_g$; Afterwards, learn the entity embedding $E_l^{new}$ after removing the forget set in the local KG of each client, and pass the $E_l^{new}$ back to the central server through the federated KG framework to obtain a new central server embedding $E_g^{new}$. The following effects need to be ensured: the new embeddings of the updated central server and each client should achieve the same effect as the embedding model retrained based on the retrained set as much as possible in downstream tasks (link prediction, etc.).

## IV. THE PROPOSED FRAMEWORK

In this section, we introduce our federated unlearning framework approach based on diffusion model (FedDM) to solve the MU problem of federated KG, and present each part of the model separately. Our model structure diagram will be shown in Figure 2.

Like most FL models, our model consists of a central server and N clients. Each client is equipped with its own local KG $G^n = (h, r, t)$, where $h, t \in E^n$, $r \in R^n$. $E^n$ denotes the set of entities of $G^n$ and $R^n$ denotes the set of relationships of $G^n$. There will be the same entities but not the same relations between different local KGs. We denote the set of all local KGs as the full set $G^f$.

In each round of interaction the server randomly updates the entity embedding $E_n^k$ of the local KG for a number of clients out of N, based on the corresponding embedding $E^k$ in the central server. After that, we divide each local KG $G^n$ into a forget set $G_n^u$ and a remaining set $G_n^r$. We use the diffusion model to learn the embedding features of $G_n^u$. After the diffusion model has been trained and stabilised, we replace the original embedding $E_n^u$ with the embedding $E_n^{u-diff}$ generated by the diffusion model. At the end of the set number of training rounds, the central server collects the embeddings of each client and performs the aggregation operation to obtain the updated embedding of the central server. In the end, optimise the model based on the defined local and global forgetting goals to achieve the goal of MU.

### A. Distribution Process

The central server selects a number of local clients at a time and selects the corresponding embedding $E^k$ in the central server for updating the local KG. the embedding $E_n^k$ of the local KG is given based on the following formula:

$$E_n^k = T^k \times E^k \tag{1}$$

where $T^k$ is the transformation matrix.

After the local KG receives the updated Enk, we compute the ternary scores of the local KG based on the traditional knowledge graph embedding model, in this article we use the TransE [17] and ComplEx [18] models for the computation with the following formulas, respectively:

(1)TransE

$$S(h, r, t) = - \| \mathbf{h} + \mathbf{r} - \mathbf{t} \|, \tag{2}$$

(2)ComplEx.

$$S(h, r, t) = Re(\mathbf{h}^T diag(\mathbf{r})\mathbf{\bar{t}}) \tag{3}$$

where $\| \cdot \|$ denotes $L_1$ norm, $Re(\cdot)$ denotes the real vector component of a complex valued vector.

Existing work [19] shows that setting a separation between global and local models is beneficial to maintain mutual compatibility between local optimisation and global generalisation of the FL framework. Inspired by this, in FedDM, we bring local and global embeddings into the knowledge graph embedding model involved in Eq. 2-Eq. 4, respectively, to obtain the local score $S_l$ and global score $S_g$, and use mutual distillation to carry out the communication process between the two. For the training process of the knowledge graph embedding model, for the embedding $(\mathbf{h}_i, \mathbf{r}_i, \mathbf{t}_i)$ of local KGs, we generate negative sample representations $(\mathbf{h}_v, \mathbf{r}_v, \mathbf{t}_v)$ of each local KG, and optimise them according to the following loss function:

$$L_{(\mathbf{h}_i, \mathbf{r}_i, \mathbf{t}_i)} = - \log \left( \sigma(S_{(\mathbf{h}_i, \mathbf{r}_i, \mathbf{t}_i)}^l) \right) - \sum \frac{1}{n} \log \left( \sigma(-S_{(\mathbf{h}_v, \mathbf{r}_v, \mathbf{t}_v)}^l) \right) \tag{4}$$

where $\sigma$ is the sigmoid activation function.

As for the distillation communication process between local embedding and global embedding, we refer to [8] and follow the following formula:

$$L_{(\mathbf{h}_i, \mathbf{r}_i)}^{distill} = KL \left( P_{(\mathbf{h}_i, \mathbf{r}_i, \cdot)}^l, P_{(\mathbf{h}_i, \mathbf{r}_i, \cdot)}^g \right) \tag{5}$$

$$P_{((\mathbf{h}_i, \mathbf{r}_i, \mathbf{t}_i)} = \frac{\exp(S_{(\mathbf{h}_i, \mathbf{r}_i, \mathbf{t}_i)}^l)}{\sum_{(h, r, t) \in (\mathbf{h}_v, \mathbf{r}_v, \mathbf{t}_v) \square \bar{\ominus}(\mathbf{h}_i, \mathbf{r}_i, \mathbf{t}_i)} \exp(S(h, r, t))} \tag{6}$$

where $KL(\cdot)$ is the Kullback-Leiber distillation function.

### B. Unlearning Process

We divide each local KG into a forget set $G_n^u = (h_i^u, r_i^u, t_i^u)$ and a remaining set $G_n^r = (h_i^r, r_i^r, t_i^r)$. We denote the embedding corresponding to the forget set as $(\mathbf{h}_i^u, \mathbf{r}_i^u, \mathbf{t}_i^u)$, and the embedding corresponding to the remaining set as $(\mathbf{h}_i^r, \mathbf{r}_i^r, \mathbf{t}_i^r)$.

Next, we employ a diffusion model to learn the embedding representation of the forgot set.

**Forward diffusion process.** The forward diffusion process is a continuous noise addition process, calculated as follows:

$$\mathbf{x}_t = \sqrt{\alpha_t}\mathbf{x}_{t-1} + \sqrt{1 - \alpha_t}\mathbf{z}_1 \tag{7}$$

$$\alpha_t = 1 - \beta_t \tag{8}$$

where $\beta_t$ represents the variance schedule across the diffusion steps [20]. t is the number of steps, $\mathbf{z}_1$ is the noise sampled

                                                                4

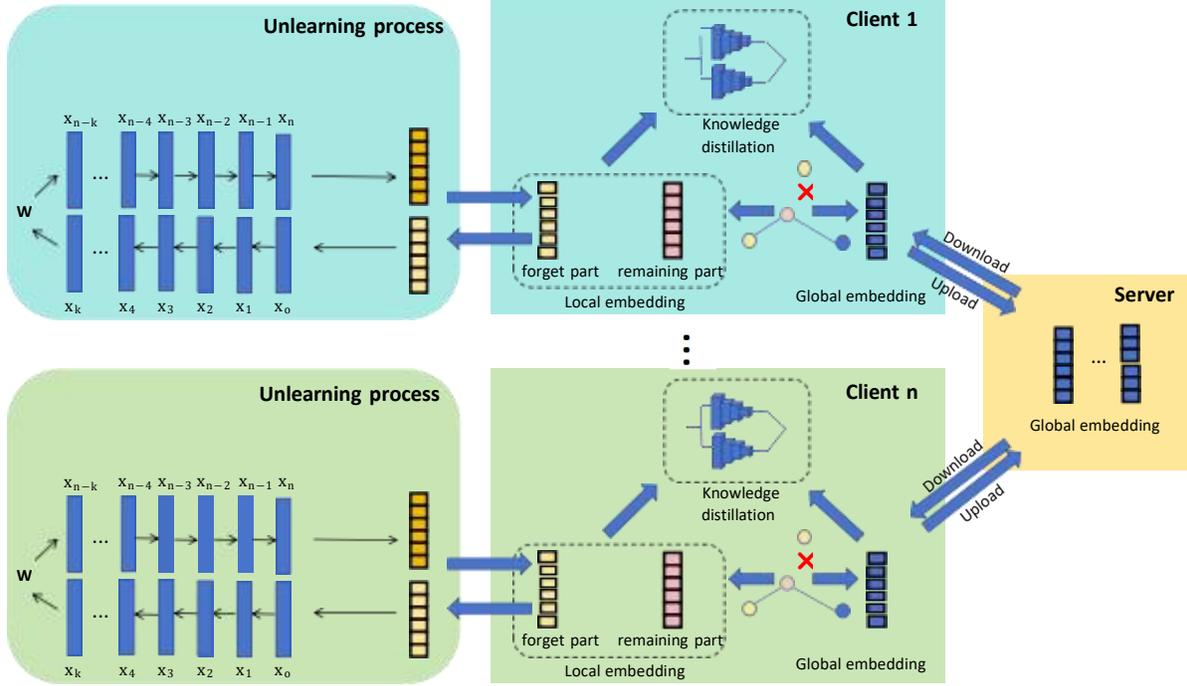

Fig. 2: The framework of FedDM.

from the standard normal distribution, $x_{t-1}$ is the representation in the previous step, and $x_0$ in the first step is initialized with the $(h_i^u, r_i^u, t_i^u)$.

After obtaining the representations in the last step $T$, $x_T$, we use the reparametrization trick to sample $w$:

$$w = \mu + \sigma \circ z_0 \qquad (9)$$

where $\mu = x_T W_\mu$ and $\sigma = x_T W_\sigma$. $W_\mu$ and $W_\sigma$ are two learnable matrices, and $z_0$ is a noise that fits the standard normal distribution.

**Reverse diffusion process.** Then we conduct the reverse diffusion, which progressively removes noise, starting with a random noise and gradually reducing to a representation without noise:

$$x_{t-1} = \frac{1}{\sqrt{\alpha_t}} \textcircled{1}_t - \frac{\sqrt{1-\alpha_t}}{\sqrt{\alpha_t}} f_\tau(\textcircled{1}_t, t) + z_2 \qquad (10)$$

where $\alpha_t$ is calculated using Equation 8, $z_2$ is also a noise that fits the standard normal distribution, $f_\tau(\textcircled{1}_t, t)$ is a neural network to approximate the conditional probabilities, which is implemented by multilayer perceptron, and $\tau$ refers to the trainable parameter in the network.

**Diffusion loss.** The diffusion process aims to train $f_\tau$ so that its predicted noise is similar to the real noise used for destruction. Following existing literature [20], the diffusion loss $L_{diff}$ is:

$$L_{diff} = E_{t \sim [0-T], f \sim \mathbb{N}(0, I)} \left[ \| f - f_\tau(x_t, t) \|^2 \right] \qquad (11)$$

where $f$ is sampled from a standard Gaussian distribution, $t$ denotes the time step, and $\| \cdot \|$ denotes the L2 distance.

After the diffusion model has been trained and stabilised, we input $(h_i^u, r_i^u, t_i^u)$ as $x_0$ to the trained diffusion model, and then replace its original representation $(h_i^u, r_i^u, t_i^u)$ with the embedding $x_n = (h_i^{u-diff}, r_i^{u-diff}, t_i^{u-diff})$ obtained in the final step n as the new representation of the forget set.

The noise addition process of the diffusion model facilitates the fading of a part of the features of the forget set, which makes the FL model adjust the corresponding parameters when learning, thus realising the purpose of machine unlearning; moreover, after the training is stabilised, the diffusion model can generate representations similar to the original embedding representations of the forget set, which retains the original forget dataset and the same overall features as the remaining set, so as to make the model's overall training volume not to decline, thereby maintaining the performance of the overall structure of the FL to be stable and not to decline drastically.

### C. Recycling Process

The next step in our model is to collect the local KGs from each client in machine unlearning after the embedding to the central server. According to what was mentioned in the previous section, the updated local KG embeddings are a new set consisting of $(h_i^r, r_i^r, t_i^r)$ and $(h_i^{u-diff}, r_i^{u-diff}, t_i^{u-diff})$. We return the entity embedding $E^l$ involved in it to the central server. Specifically, we refer to the way mentioned in a previous work [21], and obtain the updated global embedding $E^g$ by averaging the following formula:

$$E^g \leftarrow \frac{1}{n} \sum_{n E_l} \qquad (12)$$

## V. EXPERIMENT

### A. Dataset

To evaluate the effectiveness of FedDM, we select the benchmark KG embedding dataset FB15k-237 as the original dataset. Following the experiment setting of previous work, we averagely partition relations and distribute triplets into three clients accordingly, forming three datasets called FB15k-237-R3.

### B. Experimental Setup

Implementation environments. We implement FedDM with Python 3.10.0, and train the model with an NVIDIA GeForce RTX 3090 GPU with a maximum of 24GB RAM. The experiments are conducted using an Intel x86 CPU.

Evaluation Metrics. The indexes of FedDM model shown in this article are the results of training the model independently for 10 times, and then averaging. Hits@n and MRR are used as the evaluation index of the experiment to evaluate the FedDM model and the comparative models in its experiment. Hits@n indicates the average proportion of knowledge ranked less than n in link prediction, and MRR represents the reciprocal of the average ranking. The evaluation indicators used in the article are that the higher the value, the better the performance of the model.

The index calculation methods mentioned in the above indicators are defined as follows:

$$\text{Hits@n} = \frac{1}{|s|} \sum_{i=1}^{|s|} \text{indicator}(\text{rank}_i \leq n) \quad (13)$$

$$\text{MRR} = \frac{1}{|s|} \sum_{i=1}^{|s|} \frac{1}{rank_i} \quad (14)$$

where the function of indicator is: if the condition is true, the function value is 1, otherwise it is 0, $|S|$ is the number of quadruple sets, $\text{rank}_i$ refers to the link prediction ranking of the i-th quartile.

### C. Baseline

MU is an emerging task that has attracted wide attention in a short time. As mentioned in Section 2, at present, many MU methods with superior performance have been proposed. To assess the federated unlearning in FedDM, we compare the performance of raw, re-trained and unlearned embeddings. The raw embeddings are derived from federated learning of FedDM. The re-trained embeddings are re-trained from scratch using the remaining triplets. The unlearned embeddings are generated from raw embeddings through unlearning in FedDM.

### D. Results and Analysis

In Table 1, we show the MU results of FB15k-237-C3 dataset. We can draw the following conclusion: (1) The Hits@1 and MRR scores of retained model are very close, which indicates that it is not advisable to achieve MU by retraining the model. (2) The scores of unlearned model and retrained model are lower than those of the raw model, indicating that FedDM can completely eliminate knowledge through the diffusion model process, suppress memory activation on the forgot set, and thus achieve the goal of MU. (3) Unlearned model on the test set exhibit comparable performance to the raw model, which is higher than the retrained model. This indicates that FedDM can still maintain performance in downstream tasks after the process of MU.

## VI. CONCLUSION

We propose FedDM, a new diffusion model-based framework for MU of federated KG embeddings. The framework utilizes the noise property of the diffusion model to dilute the representation features of the data that need to be forgotten, which can eliminate the impact of the knowledge that needs to be forgotten on the model as a whole while still maintaining the overall performance of the federated KG embedding framework. In future work, we will explore the application of machine unlearning in downstream tasks based on the fusion of multi-source knowledge graphs. In addition, other new ideas for implementing machine unlearning in KG-related tasks will also be explored.

TABLE I: Summary of the results of each model in the experiment

| Model | TransE | | ComplEx | |
|---|---|---|---|---|
| | MRR | Hits@1 | MRR | Hits@1 |
| Raw local /forget | 59.68% | 42.17% | 60.19% | 48.17% |
| Raw local /test | 27.85% | 13.49% | 30.60% | 21.13% |
| Retrained local /forget | 17.51% | 2.34% | 21.99% | 12.06% |
| Retrained local /test | 19.58% | 6.02% | 20.38% | 13.06% |
| Unlearned local /forget | 15.88% | 1.65% | 15.33% | 6.26% |
| Unlearned local /test | 25.90% | 13.79% | 30.40% | 20.70% |
| Raw global /forget | 57.86% | 39.85% | 56.40% | 43.90% |
| Raw global /test | 27.14% | 13.01% | 28.09% | 19.02% |
| Retrained global /forget | 25.20% | 9.54% | 28.38% | 17.22% |
| Retrained global/test | 25.23% | 13.37% | 25.84% | 17.50% |
| Unlearned global /forget | 13.66% | 0.68% | 12.78% | 4.35% |
| Unlearned global /test | 25.71% | 13.45% | 29.56% | 19.93% |